\documentclass{article}

\usepackage{graphicx}
\usepackage{svg}
\usepackage{amsmath}
\usepackage{multirow}
\usepackage{float}
\usepackage{subfig}
\usepackage{array}

\usepackage{hyperref}

\usepackage[accepted]{icml2021}

\icmltitlerunning{Towards Privacy-preserving Explanations in Medical Image Analysis}

\begin{document}

\twocolumn[
\icmltitle{Towards Privacy-preserving Explanations in Medical Image Analysis}



\begin{icmlauthorlist}
\icmlauthor{Helena Montenegro}{to,goo}
\icmlauthor{Wilson Silva}{to,goo}
\icmlauthor{Jaime S. Cardoso}{to,goo}
\end{icmlauthorlist}

\icmlaffiliation{to}{University of Porto, Porto, Portugal}
\icmlaffiliation{goo}{INESC TEC, Porto, Portugal}

\icmlcorrespondingauthor{Helena Montenegro}{up201604184@up.pt}

\icmlkeywords{Deep Learning, ICML, Case-based Interpretability, Privacy}

\vskip 0.3in
]



\printAffiliationsAndNotice{}  % leave blank if no need to mention equal contribution

\begin{abstract}
The use of Deep Learning in the medical field is hindered by the lack of interpretability. Case-based interpretability strategies can provide intuitive explanations for deep learning models' decisions, thus, enhancing trust.
However, the resulting explanations threaten patient privacy, motivating the development of privacy-preserving methods compatible with the specifics of medical data. In this work, we analyze existing privacy-preserving methods and their respective capacity to anonymize medical data while preserving disease-related semantic features. We find that the PPRL-VGAN deep learning method was the best at preserving the disease-related semantic features while guaranteeing a high level of privacy among the compared state-of-the-art methods. Nevertheless, we emphasize the need to improve privacy-preserving methods for medical imaging, as we identified relevant drawbacks in all existing privacy-preserving approaches.
\end{abstract}

{\let\clearpage\relax
\section{INTRODUCTION}

Deep learning has started to prove its potential as a powerful tool to aid medical diagnosis, having achieved outstanding results in medical image analysis tasks. For instance, McKinney \textit{et al.} \cite{McKinney_cancer} developed a model for breast cancer screening whose predictions surpass those of human experts. Although deep learning cannot replace medical specialists, it can provide additional insights to support their decisions, especially when dealing with ambiguous diagnostic cases.

However, there is a reluctance to integrate deep learning models into the medical practice due to a lack of interpretability. To combat this issue, various methods have been developed in the scientific community to provide explanations for models’ decisions. 
%One type of explanations of interest is 
Case-based explanations \cite{Angelov_xDNN, Chen_ProtoPNet, Li2018DeepLF, Papernot2018deepKNN, Wilson_ComplementaryExplanations, Wilson_Ensemble,WilsonMICCAI} explain decisions about a patient by showing examples of similar cases from other patients.
Nonetheless, these explanations threaten the privacy of patients. %Therefore, there is a need to protect the patients’ identity from being leaked by privatizing the images before providing them as explanations.
Therefore, the images must be privatized before being provided as explanations.

Current privacy-preserving methods for visual data can be divided into traditional and deep learning methods.
Traditional methods are applied to the whole input, requiring an additional pre-processing step to identify the image parts that need to be privatized. Deep learning methods learn to identify the images’ sensitive regions automatically.
Regarding traditional approaches, which are applied to the whole input image, the most well-known methods are the application of filters, such as blurring \cite{Frome_Google, Neustaedter2003BalancingPA}, and the K-Same-based methods \cite{Newton_KSame, Gross_KSameSelect}. %Deep learning methods disentangle identity features from the remaining image features through an identity recognition network whose loss is backpropagated to a generative network \cite{Cho_CLEANIR, Gong_Disentangled, Oleszkiewicz_Siamese, chen2018vganbased, Wu_PP_GAN}.
Deep learning methods use an identity recognition network to guide the image privatization process. The identity recognition model can be used to perform explicit feature disentanglement, allowing to obtain identity vectors that can be directly modified during privatization \cite{Cho_CLEANIR, Gong_Disentangled}, or to create privatized images that do not possess the same identity as the original image \cite{Oleszkiewicz_Siamese, chen2018vganbased, Wu_PP_GAN}.

Although interpretability and privacy are both trending topics in current research, the fusion of case-based interpretability and privacy protection has not yet been addressed in the research community. Furthermore, only one of the referenced deep learning privacy-preserving works \cite{chen2018vganbased} considers the preservation of task-related features, which is essential in the domain of privacy-preserving case-based interpretability. In this work, we reflect on the weaknesses of current privacy-preserving approaches when applied to case-based explanations in the medical scene. Specifically, we compare three privacy-preserving methods: blurring, K-Same-Select \cite{Gross_KSameSelect}, and the deep learning model PPRL-VGAN \cite{chen2018vganbased}. 

%The requirement that the privatization process must fulfill to enable using medical images as explanations is the preservation of explanatory evidence (i.e., disease-related semantic features). %One privacy-preserving deep learning model achieves this by using a task-related classification model to guide the learning process, preserving semantic features relevant for the model to achieve its decisions \cite{chen2018vganbased}.

Furthermore, these methods are compared with respect to three requirements of privatized case-based explanations: intelligibility, privacy and explanatory evidence. Intelligibility is necessary for a human to understand the privatized image. Privacy is needed to enable using case-based explanations in the medical scene. Explanatory evidence refers to the existence of disease-related features that expose the original medical image's pathology, needed for the consumer of the explanation to understand the resemblance of the explanation to the original case.

%Although interpretability and privacy-preserving image generation are both trending topics in current research, the fusion of case-based interpretability and privacy protection has not yet been addressed in the research community. % moved upwards
%In this work, we consider some existing privacy-protection methods, which were developed mainly for biometrics or facial emotion recognition, and apply them to medical data in order to analyze their advantages and limitations in a medical scenario. We aim to verify whether current privacy-protection techniques are suitable to generate privacy-preserving explanations. This work serves as a first step towards the integration of visual privacy into explainable machine learning in medicine.

This work's main contribution is a study on the application of current privacy-preserving techniques to case-based explanations in a clinical setting, which serves as a basis for future work in the area of privacy-preserving case-based interpretability. With this work, we intend to initiate a discussion on the need to privatize visual case-based explanations to apply them to the medical scene, which has not yet been addressed in the research community.
As such, this work serves as a first step towards the integration of visual privacy into explainable machine learning in medicine.

\section{MATERIALS AND METHODS}

\subsection{Data}

The experiments use the Warsaw-BioBase-Disease-Iris v2.1 database \cite{Trokielewicz_WBDI, Trokielewicz_WBDI2}, a biometric and medical dataset of iris images containing various eye pathologies and well-defined identities that facilitate patient recognition. This database is composed of 2,996 iris images from 115 different patients, with more than 20 different eye conditions.
The most predominant eye pathologies are cataract and glaucoma. The experiments only consider images taken from the device IrisGuard AD100, constituting 1,795 images.

For the purposes of this work, we focused on one pathology: glaucoma. The images were labeled according to the presence or absence of glaucoma. The data is unbalanced, as glaucoma is present in only 425 out of the 1,795 images. In a pre-processing stage, we cropped the images to remove labels in their lower corners. During data normalization, we horizontally flipped the images of the patients' right eye, and we centered the iris of the eye in the middle of the image. 
We set the images’ resolution to $64 \times 64 $. Finally, the data was split into 65\% for training, 15\% for validation, and 20\% for testing.

\subsection{Method}

We compare traditional and deep learning privacy-preserving methods regarding their capacity to anonymize medical images while preserving explanatory evidence.

First, we use blur by applying Gaussian kernels of different sizes to study how different levels of blur affect image privacy and intelligibility. Then, we use a K-Same-based method, which averages multiple images from different identities in K-sized clusters. The selected K-Same-based method is K-Same-Select \cite{Gross_KSameSelect}, chosen due to its capacity to preserve semantic features by ensuring that all images contain the same pathologies as the original image. In this experiment, we vary the variable K, corresponding to the number of identities per averaged image, and investigate the resulting trade-off between privacy, intelligibility and explanatory evidence.

\begin{figure*}[ht!]
    \centering
    \includegraphics[scale=0.225]{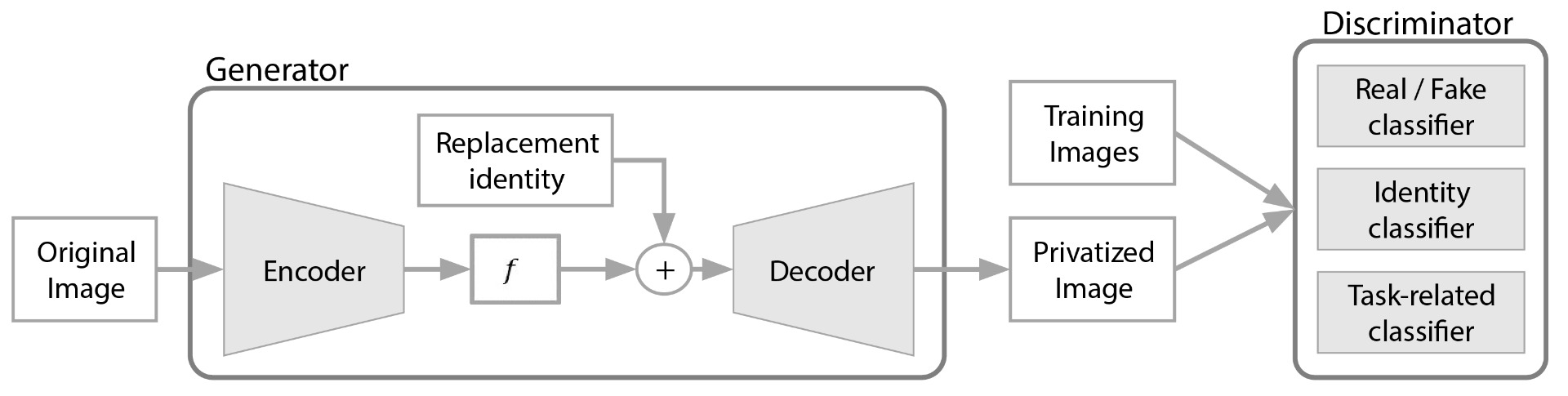}
    \caption{Overview of the PPRL-VGAN model's architecture.} \label{fig:architecture}
\end{figure*}

Finally, we use the deep learning model PPRL-VGAN \cite{chen2018vganbased}, which preserves privacy through identity replacement. This method was chosen for being the only approach that considers the preservation of semantic features needed for a particular task. In specific, this network was originally developed for privacy-preserving facial expression recognition. %PPRL-VGAN generates synthetic images that hide the original patient’s identity and preserve explanatory evidence through a Generative Adversarial Network (GAN). The GAN comprises a Variational Autoencoder (generator) that competes with a multi-task classifier (discriminator), as illustrated in Figure~\ref{fig:architecture}. The multi-task classifier contains an identity recognition network, a task-related classification network (in our case, a glaucoma recognition network), and a fake/real image classifier. The discriminator is trained to recognize identity and glaucoma in real images and distinguish between real and fake images. %On the other hand, the generator tries to trick the discriminator into classifying the synthetic image as real, minimizing its capacity to recognize fake images, and as the given replacement identity, minimizing its capacity to recognize the original identity. The model also guarantees the preservation of relevant glaucoma-related features by ensuring that the discriminator classifies the synthetic image as the original pathology.
PPRL-VGAN is a Generative Adversarial Network (GAN) comprising a conditional Variational Autoencoder (VAE) as generator and a multi-task classifier as discriminator, as shown in Figure~\ref{fig:architecture}. The generator $G$ generates an image recognized as the given replacement identity $c$, preserving the task-related features of the original image. The discriminator $D$ aids the generative task through a fake/real classifier $D^1$ to promote realism in the generated images, an identity recognition network $D^2$ to aid the identity replacement process, and a task-related classification network $D^3$ to preserve relevant semantic features. Given an image $I$ in the original data space with probability distribution $p_d$, the discriminator's loss function is shown in Equation~\ref{eq:pprl-vgan-disc-loss}, where the variables $y^{id}$ and $y^e$ correspond to the target labels for identity recognition and for the semantic task, respectively, and $\lambda^D_x$ are parameters used to calibrate the importance of each task $x$ during training.

\begin{eqnarray}
\begin{aligned}
\mathcal{L}_D= E_{(I,y,c) \sim p_d(I,y,c)} [\lambda_1^D~\log D^1(I)  + \\
 \log (1-D^1(G(I,c)))\} +  \\
 \lambda_2^D \log D_{y^{id}}^2 (I) + \lambda_3^D \log D_{y^e}^3 (I)]\label{eq:pprl-vgan-disc-loss}
\end{aligned}
\end{eqnarray}

The loss function used to train the generator is shown in Equation~\ref{eq:pprl-vgan-gen-loss}. This function includes a loss term for each of the discriminator's tasks and one term to regularize the VAE's latent space. Given an image's latent representation $f(I)$, the regularization loss term uses Kullback-Leibler Divergence (KL) to approximate the prior distribution on the latent space $p(f(I))$ and the conditional distribution $q(f(I) \mid I)$ parameterized by the encoder.

\begin{eqnarray}
\begin{aligned}
\mathcal{L}_G= E_{(I, y, c) \sim p_d(I,y,c)} [\lambda_1^G \log (1-D^1(G(I,c))) + \\ \lambda_2^G \log (1 - D_{y'(c)}^2 (G(I,c))) + \\
 \lambda_3^G \log (1 - D_{y^e}^3 (G(I,c)))] + \\ \lambda_4^G KL(q(f(I) \mid I) ~ \lvert \rvert ~ p(f(I))) \label{eq:pprl-vgan-gen-loss}
\end{aligned}
\end{eqnarray}

We compare the three methods in regards to their capacity to obtain medical case-based explanations guaranteeing intelligibility, privacy and explanatory evidence.

\subsection{Evaluation}

To evaluate the results, we use two convolutional neural networks, provided by the authors of the PPRL-VGAN model \cite{chen2018vganbased}, to classify the privatized images according to identity and presence of glaucoma. We expect to achieve poor results in identity recognition and good results in glaucoma recognition. 
These networks are trained on the original data and used as evaluation tools by examining the respective accuracy in the privatized images. The accuracy achieved by these networks on the original test set is used as the baseline. As the data is unbalanced in regards to the glaucoma-related task, we also provide the F1-Score as an evaluation metric for glaucoma recognition.
Furthermore, as some of the methods used in this work directly use images from the training data in the privatization process, we use the identity recognition network to verify whether the resulting images leak any identity from the used training images. As such, in methods where the resulting images contain an average of several images from the dataset, we check the identity recognition network's accuracy at identifying any of the used images.

Additionally, we generate visual explanations for the results of glaucoma recognition obtained in real and privatized images to understand whether the semantic features preserved using the deep learning privacy-preserving method are similar to the original image's semantic features. For this purpose, we use an implementation of the method Deep Taylor~\cite{Montavon_DeepTaylor}, provided by iNNvestigate \cite{iNNvestigate}, to visualize which image regions contribute the most to the classification.
\section{RESULTS}

The main results are summarized in Table~\ref{tab:results}, with the best results achieved for each metric and each method highlighted in bold. In this section, we describe the privatized datasets obtained with each privacy-preserving method and the results achieved with each of the evaluation networks.

\begin{table*}[h]
\centering
\caption{Experiments Results.}
\begin{tabular}{|p{0.12\linewidth}|p{0.3\linewidth}|>{\centering\arraybackslash}p{0.1\linewidth}|>{\centering\arraybackslash}p{0.115\linewidth}|>{\centering\arraybackslash}p{0.1\linewidth}|>{\centering\arraybackslash}p{0.1\linewidth}|}
\hline
\textbf{Experiment} & \textbf{Dataset} & \textbf{\begin{tabular}[c]{@{}c@{}}Identity\\ Recognition\\ Accuracy\end{tabular}} & \textbf{\begin{tabular}[c]{@{}c@{}}Replacement\\ Identity\\ Recognition\\ Accuracy\end{tabular}} & \textbf{\begin{tabular}[c]{@{}c@{}}Glaucoma\\ Recognition\\ Accuracy\end{tabular}} & \textbf{\begin{tabular}[c]{@{}c@{}}Glaucoma\\ Recognition\\ F1-Score\end{tabular}}\\ \hline
Baseline & Original test set & 90.00\% & - & 93.24\% & 87.83\% \\ \hline
\multicolumn{1}{|l|}{\multirow{9}{*}{PPRL-VGAN}}
 & Privatized set w/ random identities & \textbf{0.50\%} & 74.68\% & 79.18\% & 62.22\% \\ \cline{2-6}
\multicolumn{1}{|c|}{}
 & Privatized set w/ identities w/ the same pathologies & 1.76\% & 78.35\% & \textbf{86.56\%} & 76.01\% \\ \cline{2-6} 
\multicolumn{1}{|c|}{}
 & Privatized set w/ identities w/ different pathologies & 0.71\% & 60.26\% & 65.06\% & 48.30\% \\ \cline{2-6} 
\multicolumn{1}{|c|}{}
 & Averaged privatized set using images from random identities & 3.06\% & \textbf{13.76\%} & 85.06\% & 70.60\% \\ \cline{2-6} 
\multicolumn{1}{|c|}{}
 & Averaged privatized set using images sharing the same pathology & 2.56\% & 14.35\% & 86.24\% & \textbf{78.80\%} \\ \hline
\multirow{4}{*}{Blurring}
 & Privatized set with kernel size 3 & 69.41\% & - & \textbf{93.24\%} & \textbf{87.57\%} \\ \cline{2-6} 
 & Privatized set with kernel size 9 & 31.76\% & - & 88.82\% & 77.11\% \\ \cline{2-6} 
 & Privatized set with kernel size 15 & 23.24\% & - & 81.47\% & 55.32\% \\ \cline{2-6} 
 & Privatized set with kernel size 21 & \textbf{19.41\%} & - & 75.59\% & 32.52\% \\ \hline
\multirow{4}{*}{K-Same-Select}
 & Privatized set with 3 identities & 7.06\% & 22.94\% & \textbf{82.35\%} & \textbf{61.54\%} \\ \cline{2-6} 
 & Privatized set with 6 identities & 2.94\% & \textbf{14.41\%} & 81.76\% & 53.73\% \\ \cline{2-6} 
 & Privatized set with 9 identities & 1.47\% & \textbf{14.41\%} & 78.53\% & 42.52\% \\
 \cline{2-6} 
 & Privatized set with 12 identities & \textbf{0.88\%} & 15.29\% & 77.06\% & 32.76\% \\ \hline
\end{tabular}
\label{tab:results}
\end{table*}

%Blurring

In the experiment with blurring, we created four datasets based on the original test set, to which we applied Gaussian filters of various dimensions to achieve different degrees of blurring. In the parameter selection process, we started by selecting a small kernel size to verify the results of altering the image slightly. Then, we increased the kernel size with an interval of 6 to achieve a higher degree of privacy capable of evidencing the privacy-utility trade-off. An example of the resulting images can be seen in Figure~\ref{fig:blurring-results}. As the degree of blurring increases, the identity recognition network's capacity to identify the patient in the images decreases. Even with a high degree of blurring, the accuracy of the identity recognition network is higher than in any other method used. Additionally, higher degrees of blurring result in significant losses in the images' explanatory evidence and intelligibility, which hinders their use as explanations.

\begin{figure*}[ht!]
    \centering
    \subfloat[\centering ]{{\includegraphics[width=2.1cm]{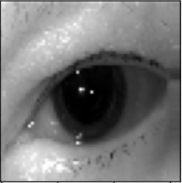} }}%
    \qquad
    \subfloat[\centering ]{{\includegraphics[width=2.1cm]{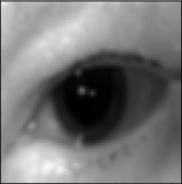} }}%
    \qquad
    \subfloat[\centering ]{{\includegraphics[width=2.1cm]{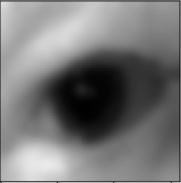} }}%
    \qquad
    \subfloat[\centering ]{{\includegraphics[width=2.1cm]{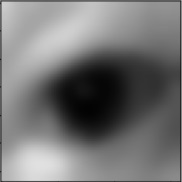} }}%
    \qquad
    \subfloat[\centering ]{{\includegraphics[width=2.1cm]{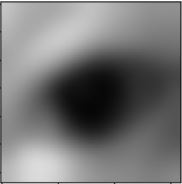} }}%
    \caption{Results of blurring the original image (a) with kernels of dimensions 3, 9, 15 and 21 (b-e).}
    \label{fig:blurring-results}%
\end{figure*}

%K-SAME

With the K-Same-Select method, we created four sets with a different number of identities used in the averaged images. Each image used in the privatized image belongs to a different patient, ensuring K-Anonymity, i.e., the highest probability of a patient being recognized is $\frac{1}{K}$ \cite{Gross_KSameSelect}. As K, we selected four values (3, 6, 9 and 12) that evidenced the trade-off between privacy and preservation of explanatory features. Figure~\ref{fig:ksame-results} shows some results of applying this technique. We achieved lower accuracy in the identity recognition network than by using blurring. As the number of identities used increases, the accuracy in both recognition networks decreases, guaranteeing higher privacy but lower explanatory value.

\begin{figure*}[ht!]
    \centering
    \subfloat[\centering ]{{\includegraphics[width=2.1cm]{images/original.jpg} }}%
    \qquad
    \subfloat[\centering ]{{\includegraphics[width=2.1cm]{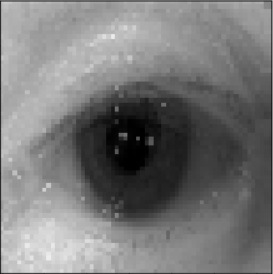} }}%
    \qquad
    \subfloat[\centering ]{{\includegraphics[width=2.1cm]{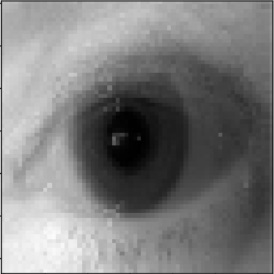} }}%
    \qquad
    \subfloat[\centering ]{{\includegraphics[width=2.1cm]{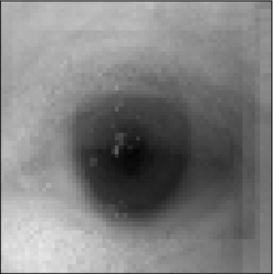} }}%
    \qquad
    \subfloat[\centering ]{{\includegraphics[width=2.1cm]{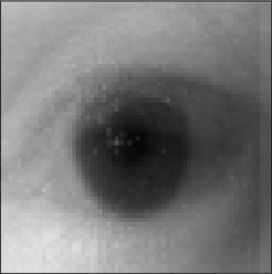} }}%
    \caption{Results of applying K-Same-Select to the input image (a) using 3, 6, 9 and 12 images from different identities (b-e).}
    \label{fig:ksame-results}%
\end{figure*}

%PPRL-VGAN

Regarding the PPRL-VGAN model, we trained the network with the hyper-parameters: $\lambda_1^G = 0.5$, $\lambda_2^G = 0.5$, $\lambda_3^G = 0.5$, and $\lambda_4^G = 0.002$, assigning the same degree of relevance to the preservation of realism, privacy, and explanatory evidence. We generated the following datasets:

\begin{itemize}
    \item Privatized set with random identities: the replacement identities were randomly selected among the patients.
    \item Privatized set with identities sharing the same pathology: the replacement identities were randomly selected from the pool of patients with the pathology observed in the input image.
    \item Privatized set with identities with a different pathology: the replacement identities were randomly selected from the pool of patients without the pathology seen in the input image.
    \item Averaged privatized set using images from random identities: six images privatized through the PPRL-VGAN network using randomly selected subjects as the replacement identities were averaged to obtain an image with a higher degree of privacy. One of the six images was privatized using the original identity.
    \item Averaged privatized set using images sharing the same pathology: six images privatized using randomly selected subjects with the same pathology as the original image were averaged. One of the six images was privatized using the original identity.
\end{itemize}

\begin{figure*}[ht!]
    \centering
    \subfloat[\centering ]{{\includegraphics[width=2.1cm]{images/original.jpg} }}%
    \qquad
    \subfloat[\centering ]{{\includegraphics[width=2.1cm]{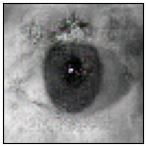} }}%
    \qquad
    \subfloat[\centering ]{{\includegraphics[width=2.1cm]{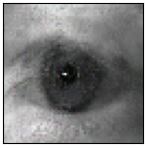} }}%
    \qquad
    \subfloat[\centering ]{{\includegraphics[width=2.1cm]{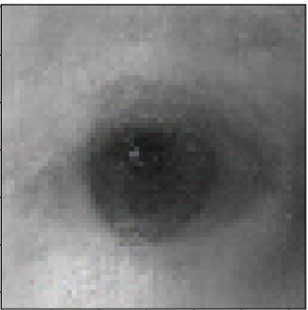} }}%
    \qquad
    \subfloat[\centering ]{{\includegraphics[width=2.1cm]{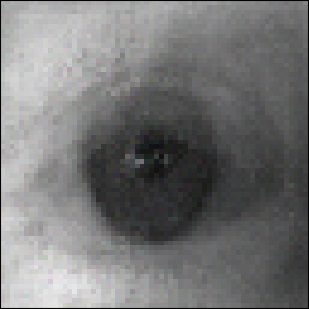} }}%
    \caption{Results of the PPRL-VGAN method. (a) represents the original image. (b) and (c) are privatized images using identities that share and that do not share the original image's pathology, respectively. (d) and (e) are images from the averaged sets considering random identities and identities sharing the same pathology as the original image, respectively.}%
    \label{fig:pprl-vgan-results}%
\end{figure*}

In this experiment, we verified that the privatized images preserve the original patient's privacy, obtaining very low accuracy when using the identity recognition evaluation model. Nonetheless, this method exposes the identity of the patients used as a replacement, threatening their privacy, as can be seen by the high accuracy when using the identity recognition network to recognize identities used as replacement in the images.

Regarding the preservation of glaucoma-related features, using random identities, we achieved low accuracy in glaucoma recognition ($\approx 79\%$), accompanied by a low F1-score ($\approx 62\%$).
However, only considering patients sharing the same pathology as the original image, we achieved higher accuracy and F1-score in glaucoma recognition, closer to the baseline. When using only patients with a different pathology, this accuracy drops to values of $\approx 65\%$. From these results, we inferred that the PPRL-VGAN model is able to reproduce disease-related features when using identities with the same pathology but struggles to do the same using identities with a different pathology. This difficulty in disentangling glaucoma-related features may be due to the nature of medical data, as most patients only possess images from one given pathology, unlike the original facial expression recognition datasets where PPRL-VGAN was validated.

%Furthermore, using the averaged privatized sets, we verified that it can protect privacy for all patients, with the restriction of K-Anonymity, and preserve glaucoma-related features to some extent. This method provided the best results, with a high glaucoma recognition accuracy and a low identity recognition accuracy.

Furthermore, we can overcome the limitation of this network in terms of the patient privacy violation for the patients used as a replacement by averaging the privatized sets. In this sense, we analyzed the averaged privatized sets, and verified that they can protect the privacy of all patients, with the restriction of K-Anonymity, and still preserve glaucoma-related features to some extent. When using privatized images that share the same identity as the original one, this method provides better results regarding preservation of explanatory evidence with slightly higher accuracy and significantly higher F1-score. Considering all the results achieved with all  privacy-preserving methods, averaging privatized images obtained through the PPRL-VGAN model is the approach that provided the best results in terms of the trade-off between privacy and explanatory evidence, with a high glaucoma recognition accuracy and a low identity recognition accuracy.

% Deep Taylor
To conclude the experiments, we use the state-of-the-art interpretability method Deep Taylor to check whether the image regions used in the diagnostic decision were the same for the privatized images obtained with PPRL-VGAN, as for the original image. Some examples are shown in Figure~\ref{fig:deep-taylor-results}. We verified that the privatization process preserves semantic features since both the original image (a) and the privatized image with glaucoma (b) display pixels with higher relevance for the classification in the same regions (upper side of the iris). Furthermore, these regions are not highlighted in the privatized image that does not contain glaucoma (c).

\begin{figure}[H]
    \centering
    \subfloat[\centering ]{{\includegraphics[width=2.1cm]{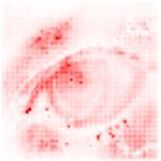} }}%
    \qquad
    \subfloat[\centering ]{{\includegraphics[width=2.1cm]{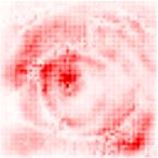} }}%
    \qquad
    \subfloat[\centering ]{{\includegraphics[width=2.1cm]{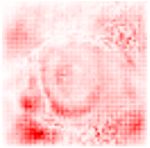} }}%
    \caption{Results of applying Deep Taylor to glaucoma recognition in: (a) input image with glaucoma, (b) privatized image with glaucoma and (c) privatized image without glaucoma.}
    \label{fig:deep-taylor-results}%
\end{figure}
\section{DISCUSSION AND CONCLUSIONS}

We studied different methods to preserve privacy in medical images, aiming to verify their potential and limitations to generate privacy-preserving case-based explanations.

In the experiments, we observed that, with blurring, to achieve a higher degree of privacy, we lose the intelligibility and explanatory evidence of the images. The significant loss of disease-related discriminative features defeats the purpose of using these privatized images as explanations.

The method K-Same-Select can preserve the images' privacy to the extent of K-Anonymity. However, as we deal with very sensitive information, K-Anonymity may not be enough to ensure privacy. In addition, while this method preserves the general features that allow classification according to a pathology, the exact features of the original image are lost.

With respect to PPRL-VGAN, we identified various issues that prevent its use in a medical scenario.
Regarding privacy, the most significant issue is the violation of the privacy of the patient identity used as a replacement.
Another issue, common to various works in the literature \cite{Cho_CLEANIR, Gong_Disentangled}, is that a multi-class classification network for identity recognition is challenging to train in the medical context due to the limited number of images per patient. To tackle this issue, we can use a Siamese network for identity recognition \cite{Oleszkiewicz_Siamese, Wu_PP_GAN}.
Regarding glaucoma recognition, when applied to medical data, this method struggled in preserving an image's disease-related class when privatizing using replacement identities from subjects that do not share the same class, failing at disentangling relevant explanatory features.
Furthermore, using a classification network does not guarantee that all relevant characteristics to diagnose glaucoma are preserved. It only ensures that the privatized image has the same classification as the original image, diminishing the resulting images' usefulness as explanations. Nonetheless, by using Deep Taylor Decomposition, we verified that the glaucoma-related semantic features are preserved during the privatization process. In the literature, many of the deep learning privacy-preserving methods \cite{Cho_CLEANIR, Gong_Disentangled, Oleszkiewicz_Siamese, Wu_PP_GAN} do not explicitly preserve task-related features, failing to guarantee the preservation of explanatory evidence.
Although we obtained the best results with the averaged set, it has the same drawbacks as the K-Same-Select method, with the limitations imposed by K-Anonymity and the loss of accurate semantic features.

Ideally, privacy protection methods should obfuscate the images' identity independently from other images in the training data, guaranteeing the privacy of all subjects. Regarding explanatory evidence, since in medical images, patients with the same disease may present different symptoms, the method should explicitly preserve the exact features in a patients' image by maximizing the similarity between these features in the original and privatized images. Also, since these images are meant to be shown to humans, their intelligibility is a fundamental requirement.
As such, none of the privacy protection methods seen in this paper nor the ones available in the literature fulfill all these requirements.

In summary, we identified the need to improve privacy protection methods for medical data. Future work will focus on developing deep learning models that are able to maintain disease-related semantic features while protecting the identity of all patients. The growing levels of realism in the images generated by state-of-the-art deep generative models~\cite{xiangli2020real,razavi2019generating,kingma2018glow} make us optimistic with regards to achieving such an ambitious goal. Having such a method would enable the use of case-based explanations, therefore, enhancing trust and, consequently, the use of deep learning to aid diagnosis. 

}

\section*{Acknowledgements}
This work was partially funded by the Project TAMI - Transparent Artificial Medical Intelligence (NORTE-01-0247-FEDER-045905) financed by ERDF - European Regional Fund through the North Portugal Regional Operational Program - NORTE 2020 and by the Portuguese Foundation for Science and Technology - FCT under the CMU - Portugal International Partnership, and also by the Portuguese Foundation for Science and Technology - FCT within PhD grant number SFRH/BD/139468/2018.


\bibliography{example_paper}
\bibliographystyle{icml2021}

\end{document}